\documentclass{article}

\PassOptionsToPackage{numbers, compress}{natbib}
\usepackage[preprint]{neurips_2023}

\usepackage[utf8]{inputenc} 
\usepackage[T1]{fontenc}    
\usepackage{url}            
\usepackage{booktabs}       
\usepackage{amsfonts}       
\usepackage{nicefrac}       
\usepackage{microtype}      

\usepackage{graphicx}
\usepackage{subfigure}
\usepackage{amsmath}
\usepackage{amssymb}
\usepackage{booktabs}
\usepackage{bbm}
\usepackage[table]{xcolor}
\usepackage[pagebackref,breaklinks,colorlinks]{hyperref}
\usepackage{pifont}
\newcommand{\xmark}{\ding{55}}%
\usepackage{bbding}
\newcommand{\rownumber}[1]{\textcolor{black}{#1}}
\usepackage{wrapfig}
\usepackage{multirow}
\usepackage{transparent}

\definecolor{brown}{RGB}{201, 104, 71}

\title{MOTRv3: Release-Fetch Supervision for End-to-End Multi-Object Tracking}

\author{
En Yu$^{1}$\thanks{This work was performed when En Yu worked as an intern at MEGVII Technology}, \ \ Tiancai Wang$^{2}$, \\ \bf{Zhuoling Li}$^{3}$, Yuang Zhang$^{4}$, Xiangyu Zhang$^{2}$, Wenbing Tao$^{1}$\thanks{Corresponding authors} \\
$^{1}$Huazhong University of Science and Technology  \\
$^2$MEGVII Technology \\
$^3$Tsinghua University  \\
$^4$Shanghai Jiao Tong University  \\
}

\begin{document}

\maketitle

\begin{abstract}

  Although end-to-end multi-object trackers like MOTR \cite{zeng2021motr} enjoy the merits of simplicity, they suffer from the conflict between detection and association seriously, resulting in unsatisfactory convergence dynamics. While MOTRv2 \cite{zhang2022motrv2} partly addresses this problem, it demands an additional detection network for assistance. In this work, we serve as the first to reveal that this conflict arises from the unfair label assignment between detect queries and track queries during training, where these detect queries recognize targets and track queries associate them. Based on this observation, we propose MOTRv3, which balances the label assignment process using the developed release-fetch supervision strategy. In this strategy, labels are first released for detection and gradually fetched back for association. Besides, another two strategies named pseudo label distillation and track group denoising are designed to further improve the supervision for detection and association. Without the assistance of an extra detection network during inference, MOTRv3 achieves impressive performance across diverse benchmarks, e.g., MOT17, DanceTrack.
\end{abstract}

\section{Introduction}

Due to its broad applications like autonomous driving and robotic navigation, multi-object tracking (MOT) is gaining increasing attention from the research community \cite{bewley2016simple,wojke2017simple}. Early MOT methods mostly adopt the \textit{tracking-by-detection} paradigm, which first recognizes targets using detection networks \cite{ge2021yolox,ren2015faster} and then associates them based on appearance similarity \cite{wang2020joint,yu2022towards,zhang2021fairmot} or box Intersection-over-Union (IoU) \cite{zhang2022bytetrack}. Although some of these methods achieve promising performance, all them demand troublesome post-processing operations, e.g., non-maximum suppression \cite{ren2015faster}.

In recent years, notable efforts have been paid to remove these post-processing operations \cite{meinhardt2021trackformer}. Among them, MOTR \cite{zeng2021motr} is a milestone, because it unifies the detection and association parts of MOT into a Transformer-based architecture and realizes end-to-end tracking without post-processing. Specifically, as shown in Fig.~\ref{fig.1} (a), MOTR first employs detect queries to recognize targets like DETR \cite{carion2020end}. When a target is located by a detect query, a track query is generated based on this detect query. The generated track query is responsible for continuously detecting this target in the following frames. Summarily, the detect queries are used to detect newly appeared targets and the track queries are for association in a implicit way. Although the MOTR architecture is elegant, it suffers from the optimization conflict between detection and association critically, which results in the poor detection precision. To alleviate this problem, significant efforts have been paid by many researchers \cite{cai2022memot, zhang2022motrv2}. For example, as illustrated in Fig.~\ref{fig.1} (b), MOTRv2 employs an independently trained 2D object detector like YOLOX \cite{ge2021yolox} to detect targets and provide the detection results to the tracking network. Then, the tracking network can concentrate on association, and thus the conflict is alleviated. Nevertheless, MOTRv2 demands a previously well-trained detector, and it makes the tracking process not in an end-to-end fashion.

\begin{figure*}[t]
\centering
\includegraphics[width=0.8\linewidth]{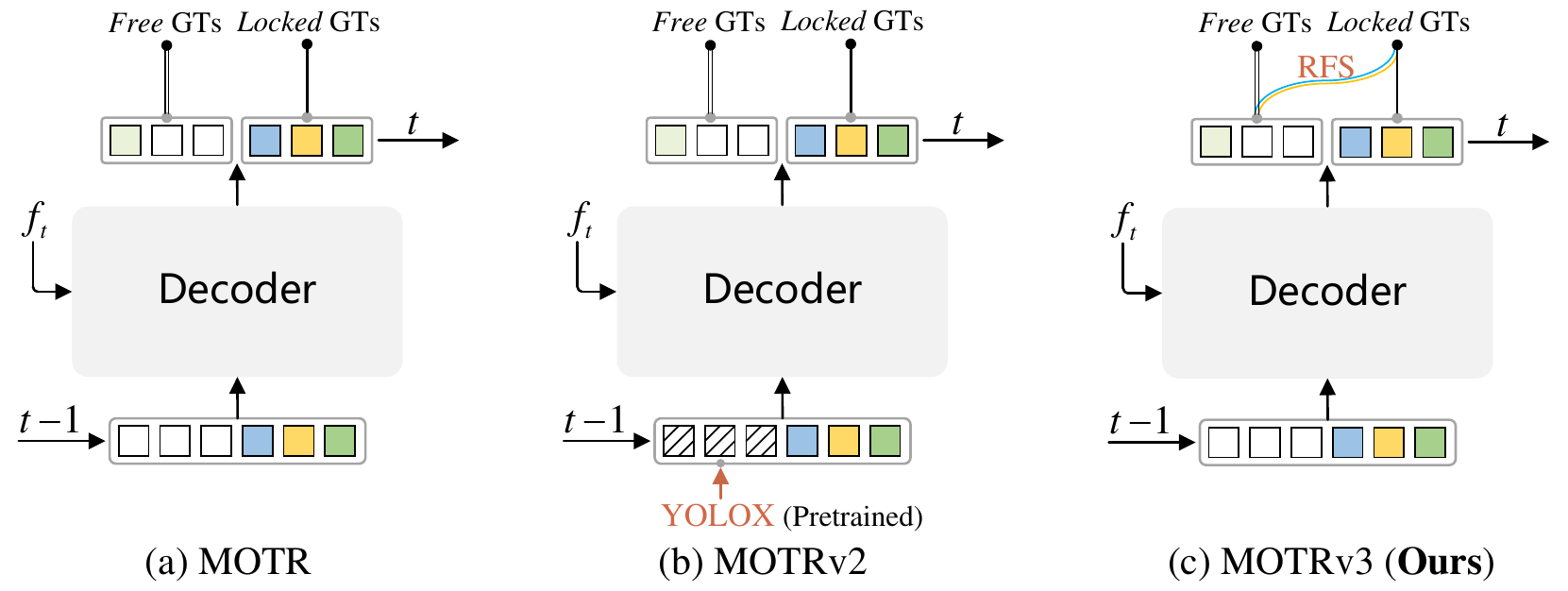}
\caption{\textbf{Comparison among MOTR, MOTRv2, and MOTRv3 (ours).} The differences in MOTRv2 and MOTRv3 compared with MOTR are marked in {\color{brown}red brown}. \textit{Locked} GTs are the labels that are assigned to track queries and \textit{free} GTs are the ones used to train detect queries.}
\label{fig.1}
\vspace{-6mm}
\end{figure*}

We argue that MOTRv2 is not elegant and does not reveal the essence of the conflict between detection and association in MOTR. In this work, we aim to explore the dark secret of the conflict and provide strategies to tackle it. To this end, we conduct numerous experiments to analyze the training dynamics of MOTR and observe that the activation times of detect queries are relatively small compared with the total number of annotated boxes. This is because when a detect query matches well with a box annotation, this box annotation will be fixedly assigned to the track query generated from that detect query. We call this annotation a locked ground truth (\textit{locked} GT). In other words, if a target appears in multiple frames, only its box label in the first frame (called \textit{free} GT) is assigned to train the detection part, and the labels in all the remaining frames are used to train the track queries. This issue causes the detection part of MOTR is not sufficiently trained. 

To tackle this issue, we propose a strategy named \textbf{R}elease-\textbf{F}etch \textbf{S}upervision (RFS), which first releases box labels to train the MOTR detection part and then automatically fetches these labels back for training the association part. Specifically, in this strategy, the one-to-one matching in MOTR detection part is conducted between all box labels and all queries (including detect queries and track queries) in the first 5 decoders, and only the matching strategy of the last decoder remains unchanged. In this way, the detection part of MOTR obtains abundant training supervision without altering the end-to-end mechanism. 

Besides, another two strategies, namely pseudo label distillation (PLD) and track group denoising (TGD), are proposed in this work to further improve the detection and association supervision, respectively. Specifically, PLD uses a previously trained 2D object detector like YOLOX \cite{ge2021yolox} or Sparse RCNN \cite{sun2021sparse} to produce pseudo labels and apply auxiliary supervision to MOTR. The distribution of pseudo labels provided by the pre-trained detector is diverse, thereby the MOTR detection part obtains more sufficient training. TGD augments track queries into multiple groups and every group consists of the same number of track queries as the original ones. Random noise is added to the reference points of each track group during training. TGD stabilizes the training of the MOTR association part and thus improves the overall tracking performance. 

Comprehensively, in this work, we reveal the underlying reason causing poor detection performance of MOTR, which previously is simply believed because of the conflict between detection and association. Based on this observation, we propose three strategies that boost the performance of MOTR by a large margin while avoiding the use of an independently trained 2D object detector like MOTRv2. Combining the developed techniques, we propose MOTRv3, which achieves impressive performances across multiple benchmarks including MOT Challenge \cite{milan2016mot16} and DanceTrack \cite{sun2022dancetrack}. We hope this work can inspire researchers about how to tackle the optimization conflicts among various subtasks.

\vspace{-4.5mm}
\section{MOTR} \label{Sec: MOTR}

MOTRv3 is implemented based on MOTR rather than MOTRv2 since it requires an extra 2D object detector, making the tracker not end-to-end. Since not all readers are clear about the design of MOTR, we first elaborate on its architecture in this section. Refer to the papers for more details \cite{zeng2021motr,zhang2022motrv2}. Afterwards, we describe how we reveal the essence resulting in the conflict between detection and association in MOTR.

\vspace{-1mm}
\subsection{MOTR Architecture} \label{SubSec: MOTR Architecture}

MOTR consists of a backbone, 6 encoders, and 6 decoders. It realizes end-to-end tracking by applying simple modifications to DETR. Specifically, when a target appears in a video, MOTR employs a detect query to recognize it in the same process as DETR. After recognizing it, MOTR uses a lightweight network block to generate a track query based on this detect query. Then, in the following frames of this video, this track query should continuously localize the positions of this target until it disappears. In a nutshell, the detect queries are utilized to detect newly appeared targets and track queries are for localizing previously detected targets.

In MOT datasets, every target in a frame is annotated with a 2D box and an identity. To enable the MOTR detection part to recognize newly appeared targets, the 2D boxes of these new targets are assigned to train detect queries during training. By contrast, if a target exists in previous frames, its box is used to train the track queries.   

\begin{figure*}[t]
    \centering
    \subfigure[MOTR]{
        \includegraphics[width=0.232\columnwidth]{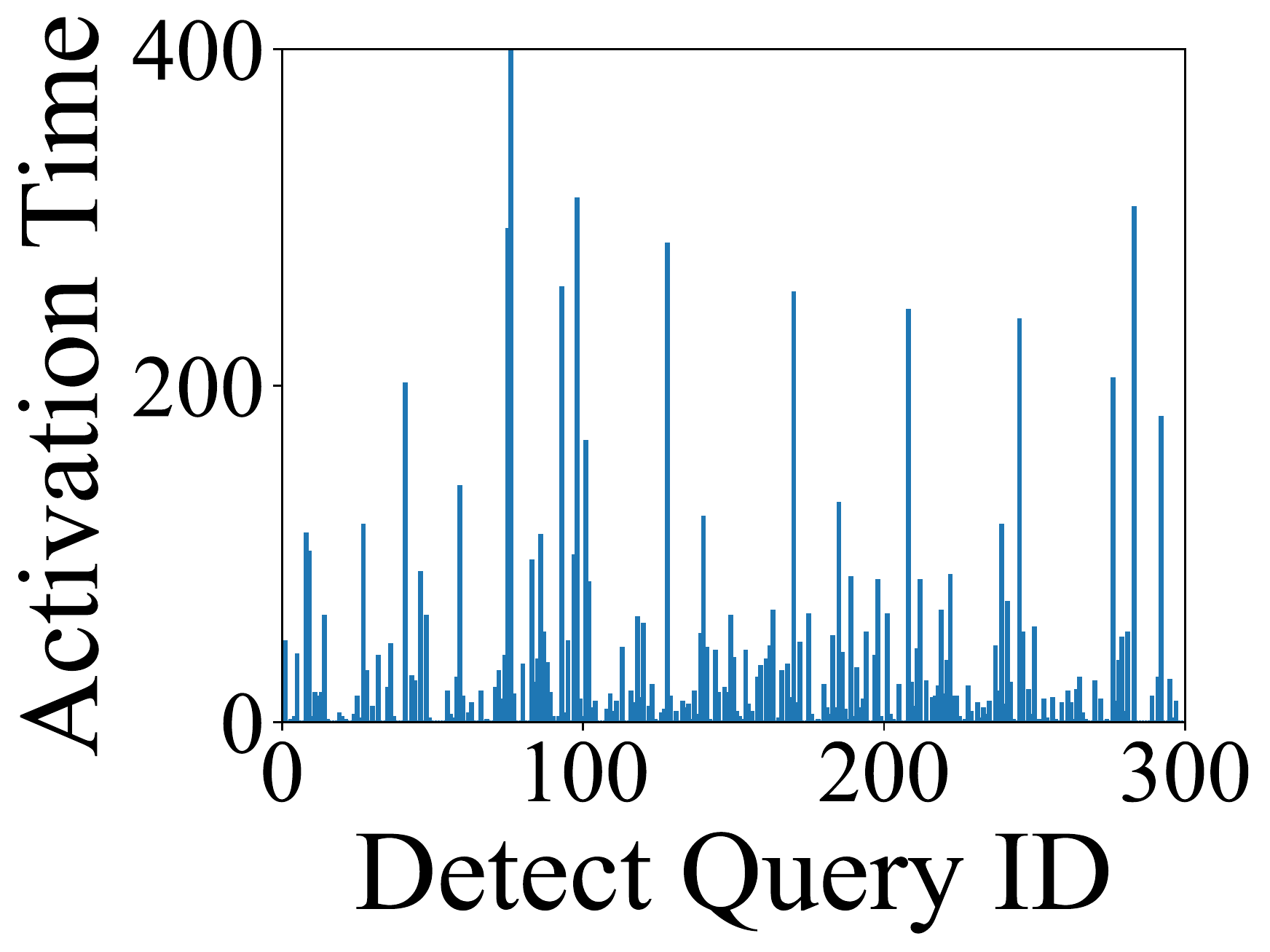}
    }
    \subfigure[MOTR with RFS]{
	\includegraphics[width=0.232\columnwidth]{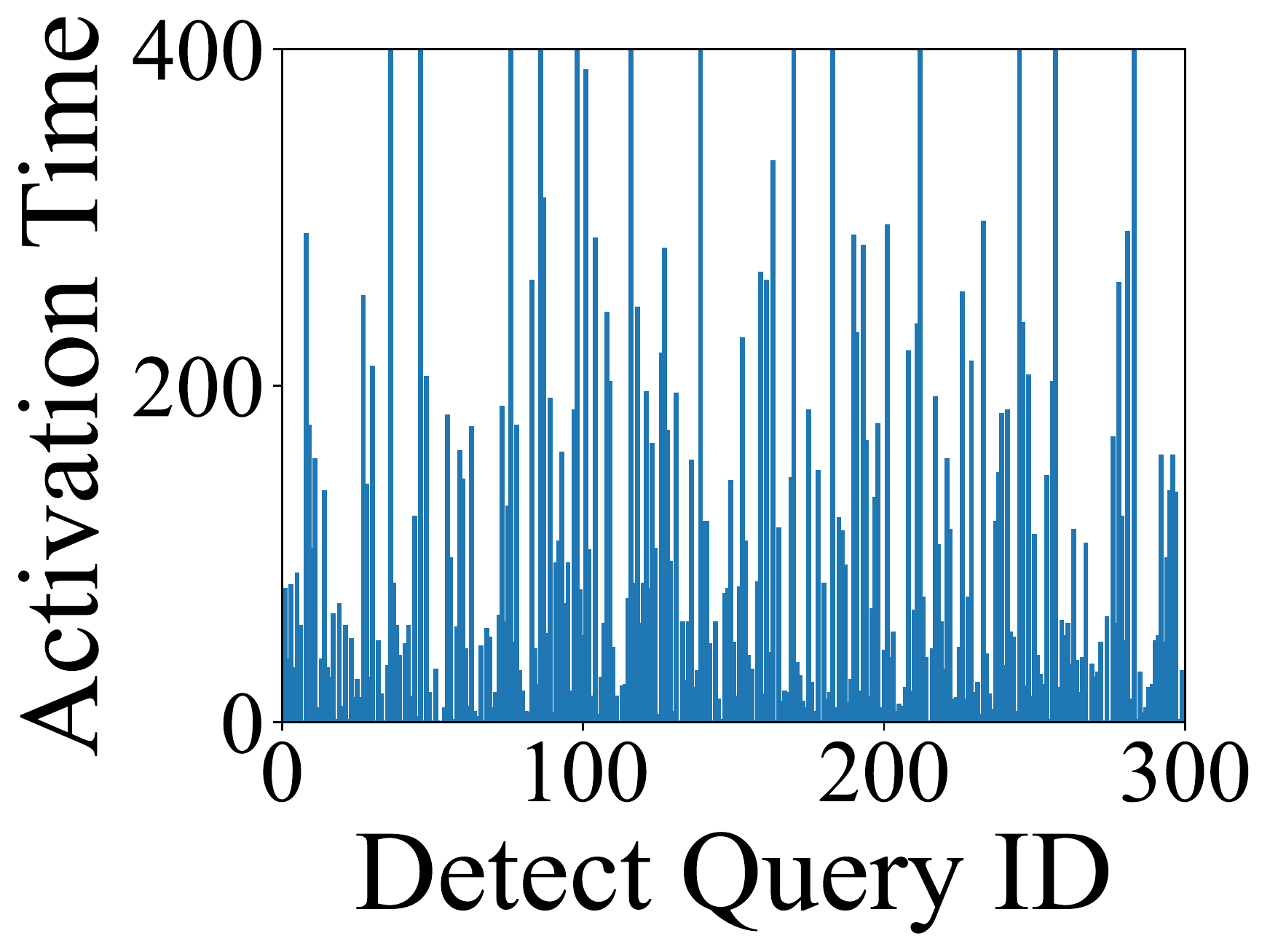}
    }
    \subfigure[MOTR]{
    	\includegraphics[width=0.232\columnwidth]{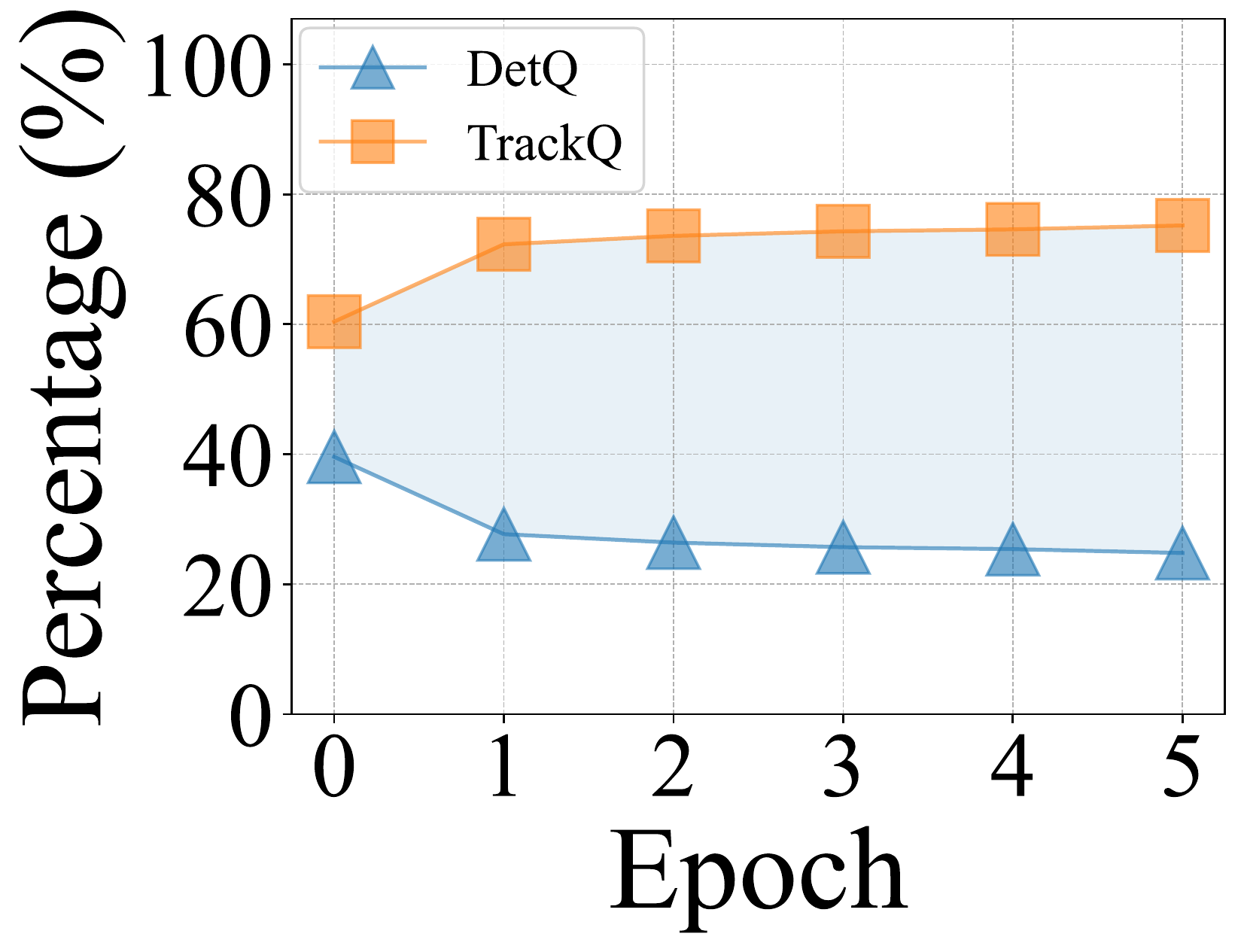}
    }
    \subfigure[MOTR with RFS]{
	\includegraphics[width=0.232\columnwidth]{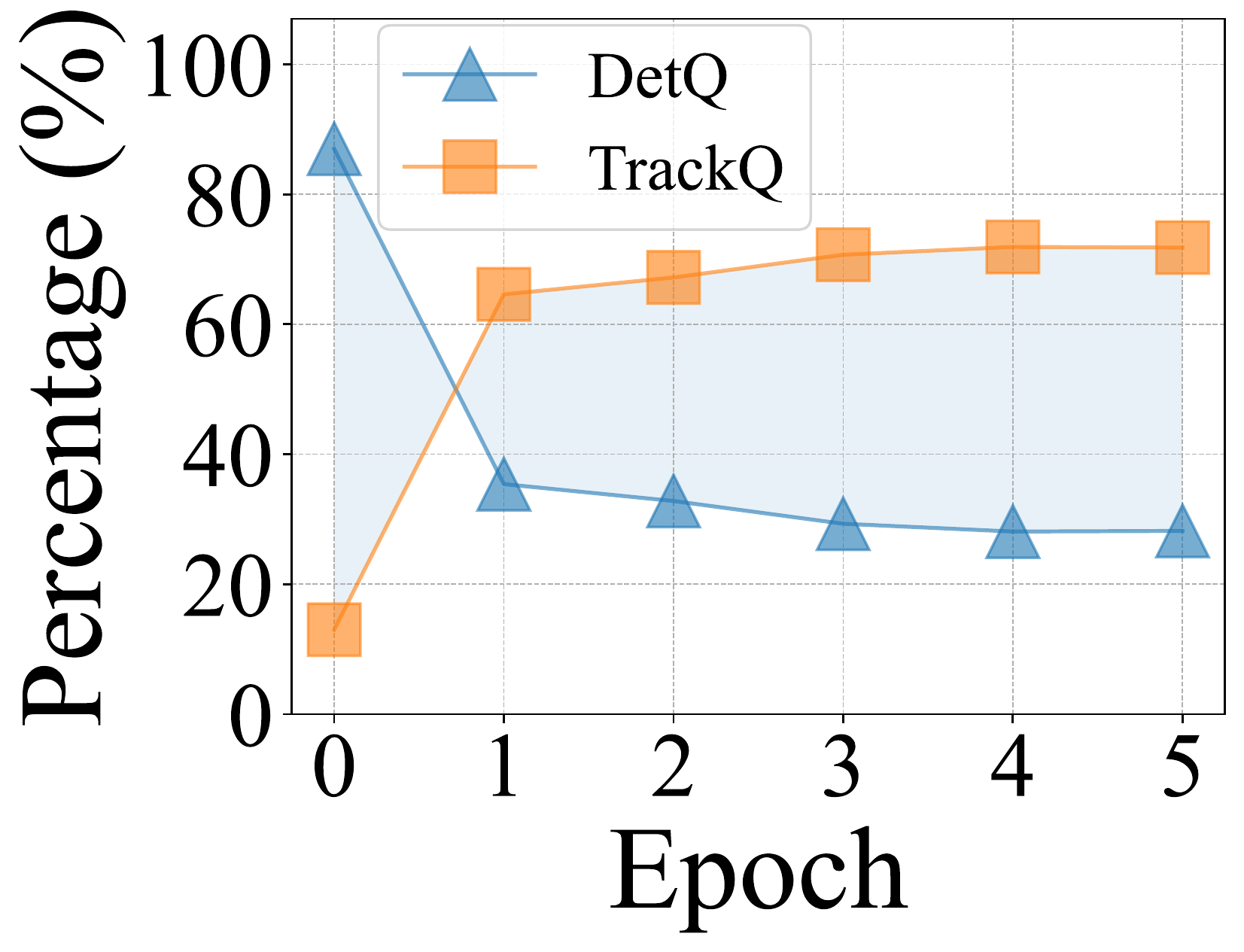}
    }
    \caption{Figure (a) and (b) show the activation number of different detect queries with and without the proposed RFS strategy during the training process. Figure (c) and (d) illustrate the dynamics of 2D box label percentages assigned to the detection and association parts in the conditions with and without RFS.}
    \label{fig.2}
\vspace{-2mm}
\end{figure*}

\subsection{A Closer Look at Label Assignment} \label{SubSec: Conflict between Detection and Association}
Although the aforementioned MOTR architecture is simple and presents promising association accuracy, its detection precision is poor. Previous literature \cite{zhang2022motrv2} commonly believes that this is due to the conflict between detection and association, but no one reveals where this conflict arises from. To shed light on this problem, we conduct an extensive analysis. As suggested in Fig.~\ref{fig.2} (a), the activation numbers of detect queries with different IDs are limited. We further compare the numbers of 2D box labels that are released to train the detect and track queries (see Fig.~\ref{fig.2} (c)). It can be observed that in the first epoch over 60\% labels are used to train the track queries while only 40\% are for the detect queries. In the following epochs, the percentage of labels assigned to track queries gradually grows, and the detect queries constantly cannot receive sufficient supervision. To alleviate this problem, we propose the RFS strategy. As shown in Fig.~\ref{fig.2} (b, d), RFS boosts the activation times and received 2D box label numbers significantly. Interestingly, although RFS enhances the percentages of labels assigned to detect queries in the initial epochs by large margins, the percentages in the final epochs are similar to the dynamics without RFS. This phenomenon implies that RFS automatically returns the labels back to the MOTR association part after the detection part is sufficiently trained. 

\section{MOTRv3}

\subsection{Overview}

\begin{figure*}[t]
\centering
\includegraphics[width=1.0\linewidth]{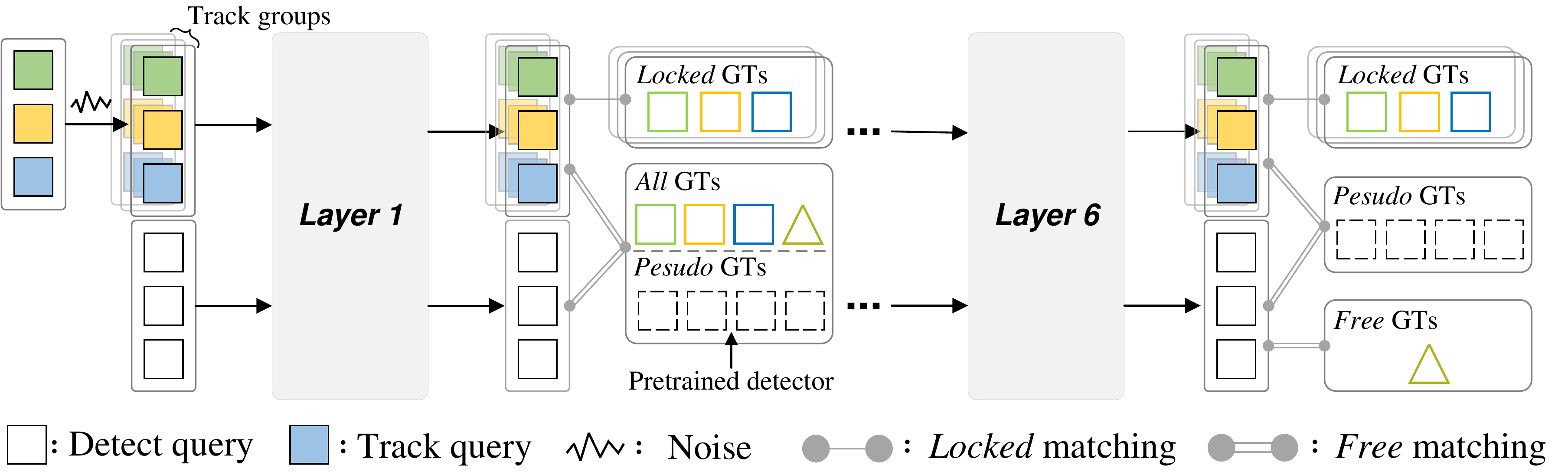}
\caption{\textbf{Overview of the MOTRv3 training pipeline}. We primarily illustrate the three proposed strategies (RFS, PLD and TGD) in this figure.}
\label{fig.3}
\vspace{-2mm}
\end{figure*}

As mentioned before, MOTRv3 is the same as MOTR except the three contributions, i.e., RFS, PLD, and TGD, which are illustrated in Fig.~\ref{fig.3}. In this section, we elaborate on the details of them one by one. Among them, RFS conducts one-to-one matching between all GTs and all queries to train the detection capability of the MOTRv3, which is different from MOTR that performs matching between only \textit{free} GTs and detect queries. As shown in Fig.~\ref{fig.2} (d), RFS releases the labels originally used for training track queries in MOTR to train the detect queries and gradually fetches them back with the progress of the training process. In PLD, a pre-trained detector is employed to produce more pseudo GTs to train the MOTR detection part more sufficiently. TGD improves the training dynamics stability of the association part by expanding track queries into several groups and then conducting the one-to-one assignment. The entire tracker is optimized with a multi-frame loss function the same as MOTR. The loss function for each frame is formulated as: $\mathcal{L} = \lambda_{cls}\mathcal{L}_{cls} + \lambda_{l_{1}}\mathcal{L}_{l_{1}} + \lambda_{giou}\mathcal{L}_{giou}$, where $\mathcal{L}_{cls}$,  $\mathcal{L}_{l_{1}}$, and $\mathcal{L}_{giou}$ are the focal loss \cite{lin2017focal}, $L_{1}$ loss and IoU loss. $\lambda_{cls}$, $\lambda_{l_1}$, $\lambda_{giou}$ are the corresponding hyper-parameters.

\subsection{Release-Fetch Supervision}

The step of matching labels with various queries for computing loss is critical for DETR-like models. For the $i_{\rm th}$ frame in a video, assume there are $K$ labels $\hat{y}^{i} = \{\hat{y}^i_j\}_{j=1}^{K}$, $M$ detect queries $q^{d}=\{q^{d}_{j}\}_{j=1}^{M}$, and $N$ track queries $q^{t}=\{q^{t}_{j}\}_{j=1}^{N}$ (usually $M+N>K$). There are two matching strategies in MOTR, one for detect queries and the other for track queries. In the first one, the labels $\hat{y}_d^i$ of newly appeared targets are assigned to detect queries $q^{d}$ based on Hungarian matching \cite{kuhn1955hungarian}. Mathematically, for the $l_{\rm th}$ decoder layer ($l = 1, ..., L$), this process is formulated as:
\begin{align}
\hat{\sigma}^{(i, l)}_{d} = \underset{\sigma^{(i, l)}_{d} \in \mathfrak{S}^{(i, l)}_d}{\arg \min } \sum_{j=1}^M \mathcal{L} \left(d^{(i,l)}_{j}, \hat{y}^{(i,l)}_{\sigma^{(i, l)}_{d}(j)}\right), \label{Eq1}
\end{align}
where $\mathfrak{S}_d$, $\sigma^{(i, l)}_{d}$, $d^{(i,l)}_{j}$, and $\mathcal{L}(\cdot)$ denote the matching space containing all possible matching combinations between $q^d$ and $\hat{y}^{i}_{d}$, a sampled matching combination, the detection result decoded from the detect query $q^{d}_{j}$, and the matching loss, respectively. $\hat{\sigma}^{(i, l)}_{d}$ represents the obtained optimal matching result. In the second matching strategy, the labels $\hat{y}_{t}^{i}$ that belong to targets appearing in previous frames are distributed to track queries with respect to the matching result in the previous frame, which is given as:
\begin{align}
\sigma^{i}_{t} = \Psi(\sigma^{i-1}_{t}, \hat{\sigma}^{(i-1, L)}_{d}), \label{Eq2}
\end{align}
where $\sigma^{i}_{t}$, and $\hat{\sigma}^{(i-1, L)}_{d}$ represent the matching result between $\hat{y}_{t}^{i}$ and $q^{t}$ in the $i_{\rm th}$ frame, and the matching pairs between $\hat{y}_{d}^{i}$ and $q^{d}$ in the final decoder layer of the $(i-1)_{\rm th}$ frame. $\Psi(\cdot)$ represents the process of generating $\sigma^{i}_{t}$ based on $\sigma^{i-1}_{t}$ and $\hat{\sigma}^{(i-1, L)}_{d}$, and this process includes operations like removing track queries corresponding to instances disappeared for continuous frames. 

When $\sigma^{i}_{t}$ is obtained as Eq.~(\ref{Eq2}), the labels matching with $d^{(i,l)}_{j}$ are removed from $\hat{y}_{d}^{i}$ and added to $\hat{y}_{t}^{i}$. 
Repeating this process in various iterations, the MOTR tracking part gradually grabs supervision labels belonging to the detection part away and causes poor detection performance.

As discussed in Section~\ref{SubSec: Conflict between Detection and Association}, the matching strategies 
allocate too few supervision labels to $q^d$. To address this problem, we propose a new matching strategy to replace the one in Eq.~(\ref{Eq1}). Specifically, the matching strategy of the final decoder layer remains unchanged. For the first $L-1$ decoder layers, we modify the detection matching space $\mathfrak{S}_d$ as $\mathfrak{S}_a$. Specifically, $\mathfrak{S}_a$ contains all possible matching combinations between all queries ($q^d$ and $q^t$) and all labels ($\hat{y}^{i}$). In this way, all labels are adopted to train both detect and track queries in the detection loss part. The labels are used to train $q^d$ or $q^t$ is determined by the similarity between labels and the boxes decoded from queries. Therefore, as illustrated in Fig.~\ref{fig.2} (d), since $q^t$ cannot precisely follow the locations of targets at the beginning of training, the labels are mostly released to train $q^d$. Then, after $q^t$ gradually be able to correctly recognize the locations of corresponding targets, the labels are fetched back to train $q^t$ automatically. Notably, we only change the detection matching strategy in RFS and the matching strategy of the association part in Eq.~(\ref{Eq2}) remains the same as MOTR.

\vspace{-3mm}
\subsection{Pseudo Label Distillation}

RFS releases more supervision labels to the detection part by changing the matching strategy. In PLD, we further enhance the supervision applied to the detection part by generating pseudo labels using a previously trained 2D object detector like YOLOX. Notably, this detector is only adopted in training and abandoned during inference, which is different from MOTRv2 that demands this detector in inference.

In PLD, we use the pretrained 2D object detector to generate detection boxes and employ a confidence threshold (such as 0.05) to select precise ones from these boxes. The selected boxes $\hat{y}^{i}_{e}$ are used as pseudo labels to train the queries of all 6 decoders. Besides the training process in RFS, we conduct one-to-one matching between all queries ($q^d$ and $q^t$) and $\hat{y}^{i}_{e}$ to compute detection loss. In this way, $q^d$ obtains more supervision.

Although the aforementioned process increases the labels for training $q^d$, the problem is that $\hat{y}^{i}_{e}$ is often noisy. To alleviate this problem, we propose to reweight the detection part loss based on the detection confidence $c_e$ produced by the 2D object detector. Specifically, if a query matches with a label, the loss is multiplied by the confidence value. If no label is matched, the query computes loss with the background class (the same as DETR) and the loss is reweighted by a factor 0.5. Mathematically, this process is formulated as:
\begin{align}
\mathcal{L}_{\sigma^{i}_{p}} = \sum\limits_{j=1}^{P} \omega_{j} \cdot \mathcal{L}(d_{j}^{i}, \tilde{y}_{\sigma^{i}_{p}(j)}^{i}), \quad \omega_{j}= 
\left\{ 
\begin{aligned} 
& c_e, \ \ \ \ if \ \tilde{y}_{\sigma^{i}_{p}(j)}^{i} \neq \varnothing  \\ 
& 0.5, \ \ else
\end{aligned}
\right \}, \label{Eq3}
\end{align}
where $\sigma^{i}_{p}$ denotes the matching results between outputs $d$ and pseudo labels $\tilde{y}$.  $P$ is the number of pseudo labels. $\omega_{j}$ and $c_e$ denote the reweighting factor and the classification score, respectively.

\vspace{-2mm}
\subsection{Track Group Denosing}

The two aforementioned strategies, RFS and PLD, improve the detection capability of MOTR. In this part, we develop a strategy, TGD, to boost the association performance. Specifically, inspired by Group DETR \cite{chen2022group}, we first augment every track query as a track query group consisting of multiple queries. Notably, the assignment between each track query group and GT is the same as the original track query. By conducting one-to-one matching between track query groups and labels and then computing loss, the track queries obtain more sufficient supervision.


Besides, we note that the tracking performance is influenced by the quality of initial reference points \cite{zhu2020deformdetr} significantly. To boost the robustness of the model, we propose to add random noise to the reference point of every element in a track query group. In this way, the model becomes less dependent on promising initial reference points and the association becomes more robust.

Furthermore, an attention mask is used to prevent information leakage \cite{li2022dn} between the original track query and the augmented queries. Mathematically, we use $\textbf{A} = [a_{ij}]_{S\times S}$ to denote the attention mask for decoders, where $S = G \cdot N + M$. Then, the values in the attention mask are defined as:
\begin{equation}
\small
a_{ij}=\left\{\begin{array}{ll}
1, & \text { if } i<M+N \text{ and } j>M+N;\\
1, & \text { if } i \geq M+N \text{ and } \lfloor\frac{i-(M+N)}{N}\rfloor \neq \lfloor\frac{j-(M+N)}{N}\rfloor;\\
0, & \text{otherwise.}
\end{array}\right.
\end{equation}
where $i$ and $j$ denote the IDs of two queries and $a_{ij}$ defines whether there should exist information communication between these two queries. After expanding the original matching space through our proposed RFS, PLD and TGD strategy, we then calculate the overall clip loss $\mathcal{L}_{clip}$ according to the matching results. Mathematically, it is formulated as:

\vspace{-5mm}
\begin{equation}
\small
\begin{split}
\hspace{-1mm}\mathcal{L}_{clip}\hspace{-0.5mm}=\hspace{-0.5mm} \sum\limits_{i=1}^{T}\hspace{-0.5mm}(\mathcal{L}_{\sigma^{i}_{r}} + \mathcal{L}_{\sigma^{i}_{p}} + \mathcal{L}_{\sigma^{i}_{g}}) / O_{i},
\vspace{-10mm}
\end{split}
\end{equation}

where $\sigma^{i}_{r}$, $\sigma^{i}_{p}$ and $\sigma^{i}_{g}$ denote the matching results in the $i^{th}$ frame obtained by RFS, PLD and TGD, respectively. The corresponding $\mathcal{L}$ represents the loss based on the matching results. $T$ is the length of the video clip and $O_{i}$ is the number of the objects in the $i^{th}$ frame.

\section{Experiments}

 \subsection{Datasets and Metrics}

 We conduct extensive experiments on three public datasets, including DanceTrack \cite{sun2022dancetrack} 
 and MOT17 \cite{milan2016mot16}, to evaluate the superiority of MOTRv3. In this part, we introduce the adopted datasets and corresponding evaluation metrics. 

 \textbf{DanceTrack}~\cite{sun2022dancetrack} is a large-scale multiple object tracking dataset with 100 video sequences in dancing scenarios. The 100 sequences are divided into 40, 25, and 35 sequences for training, validation, and testing, respectively. The targets in DanceTrack are often highly similar in appearance but present various dancing movements. This characteristic causes huge challenge to the association in MOT. In addition, the video sequences in DanceTrack are quite long (52.9 seconds on average for a sequence), which further enhances the tracking difficulty.

 \textbf{MOT17}~\cite{milan2016mot16} consists of 14 video sequences. Among them, 7 sequences are for training and the other 7 sequences are used to validate models. These sequences cover various scenarios and weather conditions, which include indoor and outdoor, day and night, etc. The targets in these video sequences are usually pedestrians moving in simple patterns, such as walking straight.


 \textbf{Metrics.} The metrics adopted in the aforementioned datasets include the HOTA \cite{hota2021} and CLEAR-MOT Metrics \cite{bernardin2008evaluating}. Specifically, HOTA consists of higher order tracking accuracy (HOTA), association accuracy score (AssA), and detection accuracy score (DetA). CLEAR-MOT Metrics include ID F1 score (IDF1), multiple object tracking accuracy (MOTA) and identity switches (IDS).

 \subsection{Experimental Settings}

 Following MOTR and MOTRv2 \cite{zeng2021motr, zhang2022motrv2, yu2022generalizing}, MOTRv3 is implemented based on Deformable-DETR \cite{zhu2020deformdetr}, which is pre-trained on COCO \cite{coco} and employs ConvNext-Base \cite{liu2022convnet} as the vision backbone. During the training process, the batch size is set to $8$. For the experiments in DanceTrack and MOT17, each batch is a video clip including 5 frames, which are selected from a video with a random sampling interval between $1$ to $10$. Following MOTRv2, track queries are generated based on detect queries when the confidences of these detect queries are above the threshold $0.5$. Adam \cite{kingma2014adam} optimizer is employed and the initial learning rate is set to $2 \times 10^{-4}$. 
 
 For the experiments in DanceTrack \cite{sun2022dancetrack}, the models are trained for $5$ epochs and the learning rate is dropped by a factor of $10$ at the $4_{\rm th}$ epoch. In MOT17 \cite{milan2016mot16}, we train models for 50 epochs and the learning rate drops at the $40_{\rm th}$ epoch.
 
 $\lambda_{cls}$, $\lambda_{l1}$ and $\lambda_{giou}$ are set to $2$, $5$ and $2$, respectively. For the implementation of PLD, the auxiliary boxes from pre-trained detectors are obtained in an offline manner. Two common 2D object detectors are adopted, which include YOLOX \cite{ge2021yolox} and Sparse RCNN \cite{sun2021sparse}. The generated 2D box predictions with confidence scores below $0.05$ are removed. In the implementation of TGD, we expand the original track query to $4$ track query groups.

\subsection{Comparison with State-of-the-art Methods}

 In this part, we compare MOTRv3 with preceding state-of-the-art methods on the two aforementioned MOT benchmarks, i.e., DanceTrack and MOT17. The results on these two benchmarks are reported in Tab.~\ref{table:dance}-\ref{table:mot17}, respectively. Without bells and whistles, MOTRv3 outperforms all compared methods in the end-to-end fashion.



\textbf{DanceTrack.} The results on the DanceTrack test set are presented in Tab.~\ref{table:dance}. As reported, MOTRv3 outperforms the baseline method MOTR \cite{zeng2021motr} by more than $16$ HOTA points on the test set ($70.4\%$ vs. $54.2\%$ HOTA). Furthermore, the tracking performance of MOTRv3 is better than MOTRv2 ($70.4\%$ vs. $69.9\%$ HOTA) without using an independent 2D object detector, which is trained on numerous extra 2D object detection data. Meanwhile, MOTRv3 achieves better detection precision than MOTRv2 according to the detection metric MOTA ($92.9\%$ vs. $91.9\%$ MOTA), which confirms the effectiveness of the proposed strategies, RFS and PLD.

\textbf{MOT17.} The experimental results on the MOT17 benchmark are shown in Tab.~\ref{table:mot17}. Similar to the results in DanceTrack, MOTRv3 outperforms MOTR by a large margin, i.e., $2.4\%$ HOTA and $4.6\%$ IDF1. Moreover, The IDS of MOTRv3 is $36.2\%$ lower than MOTR, which suggests that the obtained trajectories are continuous and robust. Compared with MOTRv2, MOTRv3 also behaves better. Furthermore, we find that the performance of MOTRv2 relies heavily on the adopted post-processing operations. If these operations are removed, the performance of MOTRv2 drops sharply, which is $57.6\%$ HOTA and $70.1\%$ MOTA. By contrast, MOTRv3 does not use any extra post-processing operations and still achieves competitive tracking accuracy. Additionally, it can be observed that ByteTrack, a CNN-based method, behaves promisingly in MOT17, although it performs inferior to MOTRv3 in DanceTrack. We infer that this is because the target movement trajectories in MOT17 are simple. Therefore, the targets in MOT17 can be tracked well by combining a strong 2D object detector like YOLOX and hand-crafted post-processing rules.


\begin{table}[t] 
\begin{center}
\caption{\textbf{Tracking results on DanceTrack \texttt{test} set.} ↑/↓ indicates that a higher/lower score is better.}

\setlength{\tabcolsep}{0.57mm}
\resizebox{0.7\columnwidth}{!}{
\begin{tabular}{l|c|ccccc}
\toprule
Method & \textit{End to end} & HOTA$\uparrow$ & AssA$\uparrow$ & DetA$\uparrow$ & MOTA$\uparrow$ & IDF1$\uparrow$ \\
\hline
\hline
\textbf{\textit{CNN-based}} &&&&&&\\
QDTrack~\cite{pang2021quasi} & \xmark & 54.2 & 36.8 & 80.1 & 87.7 & 50.4 \\
FairMOT~\cite{zhang2021fairmot} & \xmark& 59.3 & 58.0 & 60.9 & 73.7 & 72.3 \\
CenterTrack~\cite{CenterTrack} & \xmark & 41.8 & 22.6 & 78.1 & 86.8 & 35.7 \\
ByteTrack~\cite{zhang2022bytetrack} & \xmark & 47.7 & 32.1 & 71.0 & 89.6 & 53.9 \\
OC-SORT~\cite{cao2022observation} & \xmark & 55.1 & 38.3 & 80.3 & 92.0 & 54.6 \\
\midrule
\textbf{\textit{Transformer-based}} &&&&&&\\
TransTrack~\cite{sun2020transtrack} & \xmark & 45.5 & 27.5 & 75.9 & 88.4 & 45.2 \\
MOTR~\cite{zeng2021motr} & \checkmark & 54.2 & 40.2 & 73.5 & 79.7 & 51.5 \\
MOTRv2~\cite{zhang2022motrv2} & \xmark & 69.9 & 59.0 & 83.0 & 91.9 & 71.7 \\
\rowcolor{black!10} MOTRv3 & \checkmark & \bf70.4 & \bf59.3 & \bf83.8 & \bf92.9 & \bf72.3 \\
\bottomrule
\end{tabular}
}
\label{table:dance}
\end{center}
\vspace{-1mm}
\end{table}

\begin{table}[t] 
\begin{center}
\caption{\textbf{Tracking results on the MOT17 \texttt{test} set.} Notably, MOTRv2 uses extra post-processing operations for MOT17, and we remove them here for fair comparison. $*$ denotes MOTRv2 without post-processing operations.}

\setlength{\tabcolsep}{0.6mm}
\resizebox{0.75\columnwidth}{!}{
\begin{tabular}{l|c|cccccc}
\toprule
Method & \textit{End to end} & HOTA$\uparrow$ & AssA$\uparrow$ & DetA$\uparrow$ & MOTA$\uparrow$ & IDF1$\uparrow$ & IDS$\downarrow$ \\
\hline
\hline
\textbf{\textit{CNN-based}} &&&&&&&\\
QDTrack~\cite{pang2021quasi} & \xmark & 53.9 & 52.7 & 55.6 & 68.7 & 66.3 & 3,378 \\
FairMOT~\cite{zhang2021fairmot}  & \xmark& 59.3 & 58.0 & 60.9 & 73.7 & 72.3 & 3,303 \\
CenterTrack~\cite{CenterTrack} & \xmark & 52.2 & 51.0 & 53.8 & 67.8 & 64.7 & 3,039 \\
ByteTrack~\cite{zhang2022bytetrack} & \xmark & 63.1 & 62.0 & 64.5 & 80.3 & 77.3 & 2,196 \\
\midrule
\textbf{\textit{Transformer-based}} &&&&&&&\\
TransTrack~\cite{sun2020transtrack} & \xmark & 54.1 & 47.9 & 61.6 & 74.5 & 63.9 & 3,663 \\
MOTR~\cite{zeng2021motr} & \checkmark & 57.8 & 55.7 & 60.3 & 73.4 & 68.6 & 2,439 \\
\transparent{0.4}MOTRv2~\cite{zhang2022motrv2} & \transparent{0.4}\xmark & \transparent{0.4}62.0 & \transparent{0.4}60.6 & \transparent{0.4}63.8 & \transparent{0.4}78.6 & \transparent{0.4}75.0 & \transparent{0.4}- \\
MOTRv2$^*$~\cite{zhang2022motrv2} & \xmark & 57.6 & 57.5 & 58.1 & 70.1 & 70.3 & 3,225 \\
\rowcolor{black!10} MOTRv3 & \checkmark & \bf{60.2} & \bf{58.7} & \bf{62.1} & \bf{75.9} & \bf{72.4} & \bf{2,403} \\
\bottomrule
\end{tabular}
}
\label{table:mot17}
\end{center}
\vspace{-5mm}
\end{table}

\begin{table}[t] 
\begin{center}
\caption{\textbf{Overall ablation study of the proposed strategies.} The performance of the tracker employing all the developed strategies is highlighted in \colorbox{black!10}{gray}.}
\vspace{-1.5mm}

\setlength{\tabcolsep}{0.8mm}
\resizebox{0.7\columnwidth}{!}{
\begin{tabular}{@{}p{.8em}@{}lc ccc c cccccc@{}}
\toprule
 &&& \multicolumn{4}{c}{$Components$} && \multicolumn{4}{c}{$Metrics$} \\
\cmidrule{4-6}
\cmidrule{8-13}
  & Method && RFS & PLD & TGD  && HOTA$\uparrow$ & AssA$\uparrow$ & DetA$\uparrow$ & MOTA$\uparrow$ & IDF1$\uparrow$ & IDS$\downarrow$ \\
\midrule
  \rownumber{1}&	  Base &&  &  &   && 56.6 & 47.0 & 68.4 & 75.3 & 60.0 & 1,662 \\
  \rownumber{2}&	   && \Checkmark &  &   && 60.9 & 49.4 & 75.8 & 85.5 & 63.7 & 1,139 \\
  \rownumber{3}&	   &&  & \Checkmark &   && 59.2 & 46.5 & 75.5 & 84.9 & 61.7 & 1,284 \\
  \rownumber{4}&	   &&  &  & \Checkmark && 59.6 & 49.7 & 71.9 & 80.0 & 62.1 & 1,804 \\
  \rownumber{5}&	   && \Checkmark & \Checkmark &  && 61.7 & 50.0 & 76.3 & 86.0 & 64.8 & 1,350\\
  \rowcolor{black!10} \rownumber{6}&	   MOTRv3 && \Checkmark & \Checkmark &  \Checkmark && \bf 63.9 & \bf 53.5 & \bf 76.7 & \bf 86.8 & \bf 67.2 & \bf 1,151 \\
\bottomrule
\end{tabular}
}
\label{table:ablation}
\end{center}
\vspace{-2mm}
\end{table}

 \begin{table}[t]
\begin{center}
\caption{\textbf{The tracking results of using different pseudo label generation strategies.}}
\setlength{\tabcolsep}{0.8mm}
\resizebox{0.75\columnwidth}{!}{
\begin{tabular}{l |c c c c c c c }
\toprule
Pretrained detector & HOTA$\uparrow$ & AssA$\uparrow$ & DetA$\uparrow$ & MOTA$\uparrow$ & IDF1$\uparrow$ & IDS$\downarrow$ \\
\midrule

YOLOX~\cite{ge2021yolox} & 61.6 & 49.8 & 76.4 & 86.1 & 63.4 & 1,408 \\
Sparse RCNN~\cite{sun2021sparse}&  \bf 61.7 &  50.0 & 76.3 & 86.0 & \bf 64.8 & \bf 1,350 \\
Parallel (YOLOX \& Sparse RCNN) & 60.4 &  47.4 & \bf 77.1 & \bf 86.7 & 62.5 & 1,551 \\

\midrule
Ground Truth & 60.3 & 48.3 & 75.6 & 85.6 & 63.5 & 1,411\\
\bottomrule
\end{tabular}
}
\label{table:aux_sup}
\end{center}
\vspace{-5mm}
\end{table}

\subsection{Ablation Study}

 In this part, we perform extensive ablation study experiments using the DanceTrack \texttt{validation} set to analyze the effectiveness of various proposed strategies in MOTRv3. The baseline method is MOTR with anchor queries. All the models are trained using the DanceTrack training set for $5$ epochs.

 \textbf{Overall ablation study.} In this part, we study the overall influence of the three proposed strategies (RFS, PLD, and TGD) on the MOTRv3 performance. The results are reported in Tab.~\ref{table:ablation}. According to the results, all these three strategies boost the tracking performance significantly. Among these strategies, both RFS (row \#2) and PLD (row \#3) enhance the tracking precision by a large margin. Specifically, RFS improves the MOTA score by $10.2\%$ and DetA score by $7.4\%$. PLD boosts the MOTA score by $9.6\%$ and DetA score by $7.1\%$. The results indicate that both fair assignment strategy and aux supervision improve the detection capability of MOTR quite effectively. Additionally, combing them further improves the tracking performance by a large margin (row \#5). This is because RFS guarantees that a proper ratio of labels is released to train the detect queries, and PLD helps generate more detection labels. Combining them enables the MOTR detection part to be sufficiently trained.  Moreover, it can be observed that TGD improves the AssA score by $2.7\%$ and IDF1 score by $2.1\%$. This observation indicates that the representing target ability of track queries is improved, and thus the produced trajectories become more continuous and robust.

 Incorporating all these strategies, MOTRv3 (row \rownumber{6}) outperforms the baseline (row \rownumber{1}) by $7.3\%$ on HOTA and $11.5\%$ on MOTA. Summarily, the experimental results demonstrate that the proposed strategies can address the conflict between detection and association existing in end-to-end trackers effectively and MOTRv3 is an efficient end-to-end tracker. 

 \textbf{PLD.} PLD is responsible for producing more training labels. In this part, we study how different pseudo label generation strategies affect tracking performance. Specifically, we compare 4 strategies, i.e., directly copying GTs, generating pseudo labels by YOLOX or Sparse RCNN, and combining pseudo labels (concat or parallel) from YOLOX and Sparse RCNN. The results are presented in Tab.~\ref{table:aux_sup}. Two observations can be drawn. First of all, using the pseudo labels produced by detectors brings better performance than directly employing real labels. We infer that this is because the boxes produced by an extra detector are more diverse than GTs and these boxes contain detection confidence values. Secondly, employing any one of YOLOX and Sparse RCNN leads to promising performance improvement and combining them further improves the detection performance. However, the association performance tends to decrease when combining in parallel. We speculate that this is because the distributions of boxes generated by them are different and this issue confuses the learning of association during training. 

\begin{table}[h] 
\begin{center}
\small
\caption{\textbf{Ablation Study on how TGD affects the tracking performance.}}

\setlength{\tabcolsep}{0.8mm}
\resizebox{0.8\columnwidth}{!}{
\begin{tabular}{l |c |c c c c c c c }
\toprule
Method & Group Num & HOTA$\uparrow$ & AssA$\uparrow$ & DetA$\uparrow$ & MOTA$\uparrow$ & IDF1$\uparrow$ & IDS$\downarrow$ \\
\midrule
base & & 61.7 & 50.0 & 76.3 & 86.0 & 64.8 & 1,350 \\
\midrule
+ track query group & 3 & 63.2 & 52.4 & 76.6 & 86.1 & 65.9 & 1,116 \\
& 4 & 63.7 & 53.3 & \bf 76.8 & 86.7 & 66.6 & \bf 1,027 \\
& 5 &63.3 & 52.6 & 76.6 & 86.3 & 66.5 & 1,106 \\
\midrule
+ reference point noise  & 4 & \bf 63.9 & \bf 53.5 & 76.7 & \bf 86.8 & \bf 67.2 & 1,151 \\
\bottomrule
\end{tabular}
}
\label{table:tgd}
\end{center}
\vspace{-3mm}
\end{table}

\textbf{TGD.} In this experiment, we study how the track query number contained in a group affects performance and the influence of noise added to the track query reference points. The results are reported in Tab.~\ref{table:tgd}. As shown in the $1_{\rm st} \sim 4_{\rm th}$ rows of results, augmenting every query into a group improves the performance significantly and setting the query number to 4 results in the best result. Augmenting a query into too many or too few queries both harm the final tracking performance.  Besides, adding noise to reference points also boosts the tracking precision significantly, which is given in the $5_{\rm th}$ row. The results suggest that the developed TGD strategy enhances the association accuracy of MOTR significantly, which we believe is because the stability of the training process is improved.

\begin{wrapfigure}{r}{5.5cm} 
    \vspace{-0.5cm} 
    \centering
    \footnotesize
    \includegraphics[width=0.4\columnwidth]{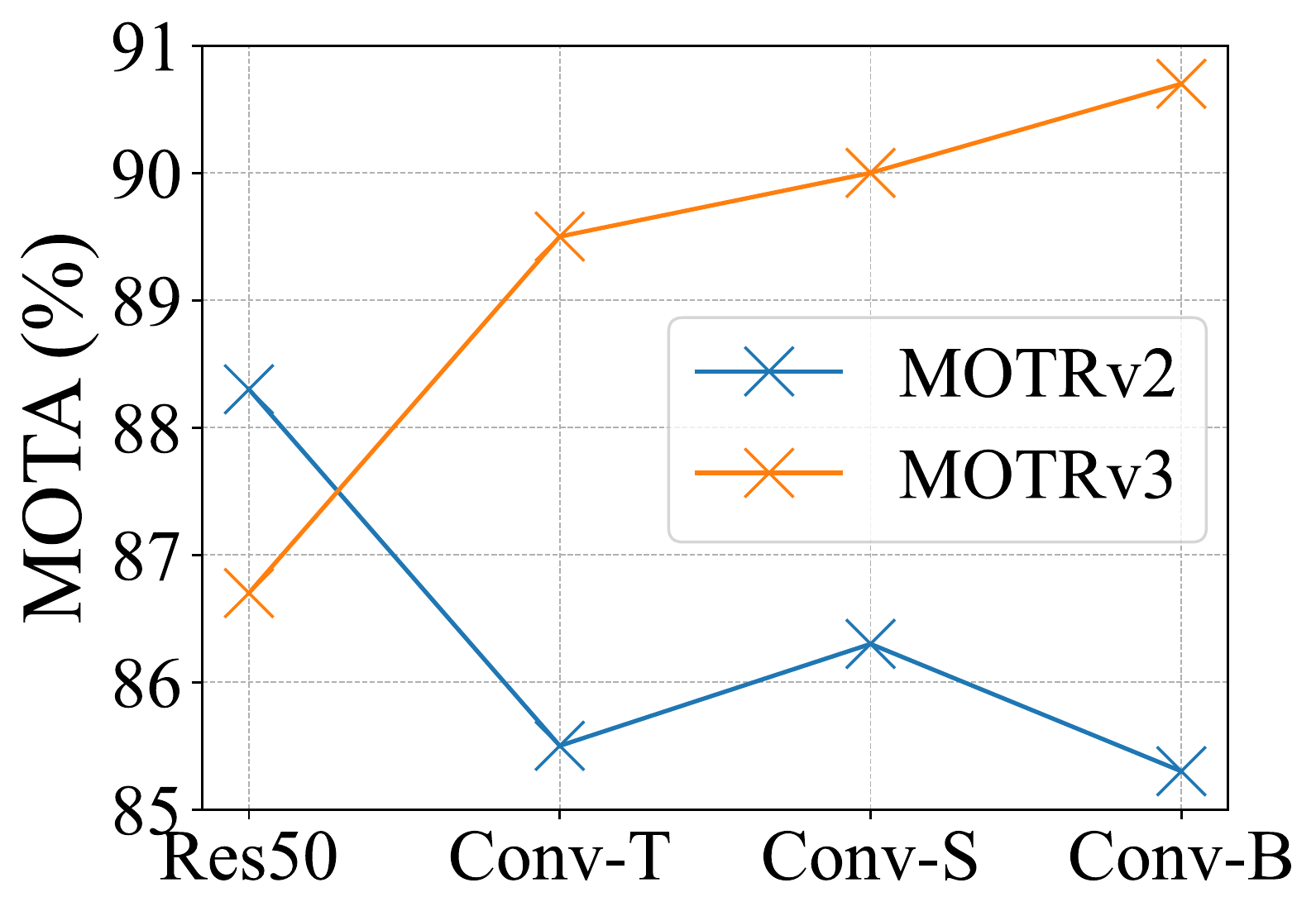}
    \vspace{-5mm} 
    {
        \caption{\textbf{Model scaling up.} Res50, Conv-T, Conv-S and Conv-B denote ResNet-50, ConvNext-Tiny, ConvNext-Small and ConvNext-Base, respectively. \vspace{-0.3cm} }
        \label{fig.5}
    }
    \vspace{-6mm} 
\end{wrapfigure}

\textbf{Model scaling up.} In this experiment, we study how scaling up backbones affects the tracking performance of MOTRv2 and MOTRv3. Specifically, we replace the original ResNet-50 backbone of them with ConvNeXt-tiny, ConvNeXt-small, and ConvNeXt-base, respectively. The results are illustrated as Fig.~\ref{fig.5}. It can be observed that the performance of MOTRv3 is continuously boosted with the scaling up of backbones. However, scaling up the backbone harms the tracking precision of MOTRv2. We speculate that this is because MOTRv2 needs an extra detector and is not end-to-end, thereby only replacing the backbone of the association part does not improve the overall tracking performance. By contrast, MOTRv3 is fully end-to-end and thus enjoys the benefits from scaling up the model.

\section{Related Works}
\textbf{Tracking by detection.} Thanks to the fast development of object detection techniques \cite{ren2015faster, zhou2019objects, ge2021yolox}, existing MOT methods mainly follow the tracking-by-detection (TBD) paradigm \cite{bewley2016simple, wojke2017simple, tracktor, peng2020chained, pang2020tubetk, wang2019towards}, which first uses detectors to localize targets and then associate them to obtain tracklets. According to the association strategy, MOT methods can be further divided into motion-based trackers and appearance-based trackers. Specifically, motion-based trackers \cite{zhang2022bytetrack, cao2022observation} perform the association step based on motion prediction algorithms, such as Kalman Filter \cite{bishop2001introduction} and optical flow \cite{baker2004lucas}. Some motion-based trackers \cite{feichtenhofer2017detect, tracktor, han2020mat, CenterTrack, sun2020transtrack, shuai2021siammot} directly predict the future tracklets or displacements in future frames compared with the current frame. In contrast to the motion-based methods, the appearance-based trackers \cite{wang2020joint, zhang2021fairmot, yu2022relationtrack, yu2022towards} usually use a Re-ID network or appearance sub-network to extract the appearance representation of targets and match them based on representation similarity.

\textbf{End-to-end MOT.} Although the performance of TBD methods is promising, they all demand troublesome post-processing operations, e.g., non-maximum suppression (NMS) \cite{ren2015faster} and box association. Recently, the Transformer architecture \cite{vaswani2017attention} originally designed for natural language processing (NLP) has been applied to computer vision. For instance, DETR \cite{carion2020end} turns 2D object detection into a set prediction problem and realizes end-to-end detection. Inspired by DETR, MOTR \cite{zeng2021motr} transfers MOT to a sequence prediction problem by representing each tracklet through a track query and dynamically updating track queries during tracking. In this way, the tracking process can be achieved in an end-to-end fashion. However, despite MOTR enjoys the merits of simplicity and elegance, it suffers the limitation of poor detection performance compared to the TBD methods. To improve MOTR, MeMOT \cite{cai2022memot} builds the short-term and long-term memory bank to capture temporal information. LTrack \cite{yu2022generalizing} introduces natural language representation obtained by CLIP \cite{radford2021learning} to generalize MOTR to unseen domains. MOTRv2 \cite{zhang2022motrv2} incorporate the YOLOX \cite{ge2021yolox} object detector to generate proposals as object anchors, providing detection prior to MOTR.

\section{Limitation and Conclusion}

In this work, we reveal the real reason causing the conflict between detection and association in MOTR, which results in the poor detection. Based on this observation, we propose RFS, which improves the detection and overall tracking performances by a large margin. However, while RFS helps mitigate this conflict in terms of supervision, the trade-off between detection and association remains unresolved. How to disentangle two sub-tasks still deserves further study. Besides, we have proposed two another strategies, PLD and TGD, to further improve the detection and query parts of MOTR. Combining all the three strategies, the developed tracker, MOTRv3, has achieved impressive performances across multiple benchmarks. We hope this work can inspire more solid works about MOT in the future.

\bibliographystyle{splncs04}
\bibliography{egbib}
\newpage


\appendix

\section{Appendix}

In this appendix, we provide more details of MOTRv3 due to the 9-page limitation on paper length. Specifically, Section~\ref{tgd} presents more details about the proposed track group denoising (TGD) strategy. Section~\ref{more imp} elaborates on the auxiliary 2D object detector employed in the pseudo label generation strategy. Section~\ref{more exp} provides additional experiments to analyze the characteristics of MOTRv3. 

\section{More Details about Track Group Denosing}
\label{tgd}

This section provides more details about the implementation of TGD. As depicted in Fig.~\ref{fig.1}, TGD initially expands a single group of track queries into $K$ groups, and then concatenates them before feeding them into the decoder. Moreover, the reference boxes of the track queries are also replicated $K$ times. Subsequently, random noise is applied to the augment reference boxes. Notably, we only scale the reference boxes (alter the width and height) rather than changing their box centers. In this way, the matching results of different track query groups remain the same as the original track queries. Thus, we do not need to recalculate the matching result.

In addition, the attention mask used for preventing information leakage\cite{li2022dn} is crucial for TGD. Specifically, there are two types of information leakage needing to be addressed, the first one is between original track queries and augmented track queries, and the other one is between different augmented track query groups. As shown in Fig.~\ref{fig.1}, the attention mask is used to address these two types of information leakage.

\begin{figure*}[h]
\centering
\includegraphics[width=0.8\linewidth]{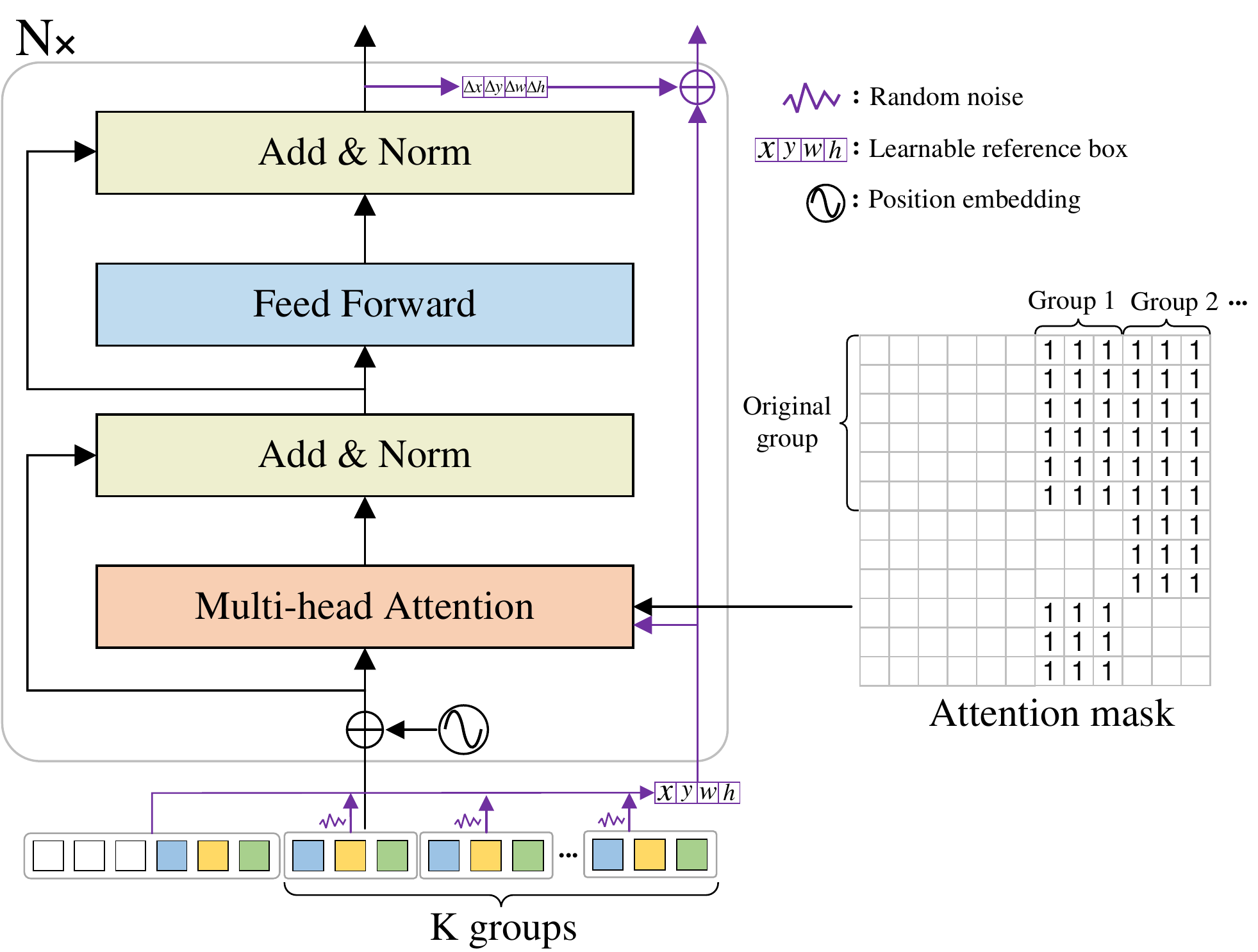}
\caption{\textbf{Illustration of the TGD strategy.} We only illustrate the process of one decoder layer for example, and the other decoders share the same procedures. First of all, the original track queries are expanded into $K$ track query groups. Subsequently, the decoder takes in all these query groups to perform one-to-one matching. Besides expanding track queries, the reference boxes of queries are also expanded as $K$ groups, and random noise is added to these reference boxes. To prevent information leakage between original track queries and the expanded track query groups, an attention mask is applied.}
\label{fig.1}
\vspace{-2mm}
\end{figure*}

\section{Auxiliary Detectors used in Pseudo Label Generation}
\label{more imp}

In this work, we mainly use YOLOX \cite{ge2021yolox} and Sparse RCNN \cite{sun2021sparse} detectors to generate pseudo labels.

\textbf{YOLOX.} We employ the YOLOX detector that the model weights are from ByteTrack \cite{zhang2022bytetrack} and DanceTrack \cite{sun2022dancetrack}. The hyper-parameters and data augmentation techniques, including Mosaic  \cite{bochkovskiy2020yolov4} and Mixup, remain consistent with ByteTrack. YOLOX-X~\cite{ge2021yolox} is adopted as the backbone. For the results on MOT17, the model is trained for $80$ epochs combining the data from MOT17, Crowdhuman, Cityperson, and ETHZ datasets. Regarding DanceTrack, we directly used the YOLOX weight provided by the DanceTrack official GitHub repository\footnote{\url{https://github.com/DanceTrack/DanceTrack}}.

\textbf{Sparse RCNN.} We utilize the original Sparse RCNN implemented in the official repository\footnote{\url{https://github.com/PeizeSun/SparseR-CNN}}, with the ResNet-50 backbone~\cite{He2016Resnet} initialized from a COCO-pretrained model. The number of learnable anchors is set to 500. To train on the MOT17 dataset, we initially train Sparse RCNN on Crowdhuman for 50 epochs. Subsequently, we further fine-tune it on MOT17 for additional 30 epochs. Similarly, for the process on DanceTrack, we also first pre-train Sparse RCNN on Crowdhuman for 50 epochs, and then fine-tune it on DanceTrack for 20 epochs.

\section{Additional Experiments.}
\label{more exp}

\begin{wrapfigure}{r}{6.0cm} 
    \vspace{-3mm} 
    \centering
    \footnotesize
    \includegraphics[width=0.4\columnwidth]{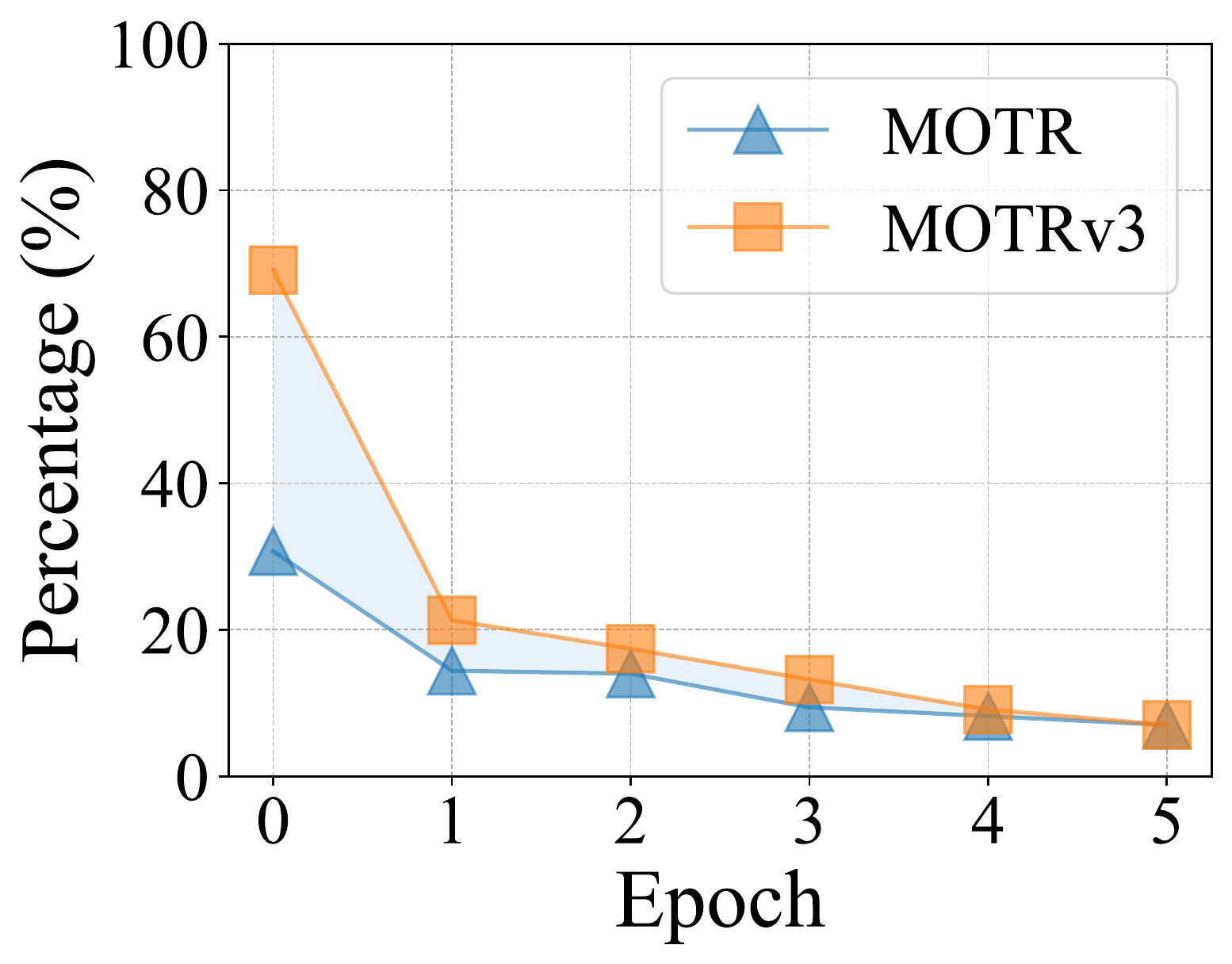}
    \vspace{-3mm}
    {
        \caption{Matching results diversity during training. \vspace{-1mm} }
        \label{fig.4}
    }
\vspace{-3mm}
\end{wrapfigure}

\textbf{Ablation study on RFS.} By applying RFS to MOTR, we allow detect queries and track queries to compete for the supervision labels fairly in the first 5 decoders. In this way, the detect queries of MOTRv3 obtain more sufficient supervision compared with the ones in MOTR. Nevertheless, RFS could result in inconsistent learned label assignment patterns between the first 5 decoders and the last decoder due to different assignment strategies during training, which may be harmful to the final tracking performance. In this part, we study this issue by visualizing the diversity between the first 5 decoder layers and the last decoder layer during training. 

Specifically, for a detect query, if the matched label is different between the last decoder layer and one of the 5 decoder layers, we call the label assignment is misaligned. In this experiment, we count the percentages of misaligned labels relative to the total labels for trackers using and without using RFS. The misalignment percentages of two trackers in various epochs are visualized in Fig.~\ref{fig.4}. The graph clearly indicates that the usage of RFS amplifies the misalignment of label matching during the initial training epochs, but over time, the percentages gradually decrease and eventually reach the same level as those without RFS. This observation suggests that the high matching diversity introduced by RFS in the early training stage does not hinder the convergence of the label matching process. In fact, the increased matching diversity allows more queries to participate in the learning process, which ultimately benefits the detection part.


\textbf{Analysis on the inference speed.} As mentioned in the main paper, our proposed strategies, namely RFS, PLD, and TGD, are exclusively employed during training and do not introduce any additional network blocks. Consequently, the inference speed of MOTRv3 remains competitive. As depicted in Tab.~\ref{table:fps}, we compare the inference speeds of MOTR, MOTRv2, and MOTRv3 on the DanceTrack \texttt{test} set. It can be observed that our MOTRv3, with the ResNet-50 backbone, achieves the highest inference speed compared to MOTR and MOTRv2. Moreover, MOTRv3 with the ConvNext-Base backbone achieves superior performance while still maintaining a competitive inference speed.

\begin{table}[h] 
\begin{center}
\caption{\textbf{Inference speed comparison on DanceTrack \texttt{test} set among MOTR series.}}

\setlength{\tabcolsep}{3.5mm}
\resizebox{0.85 \columnwidth}{!}{
\begin{tabular}{l |c | c c c | c }
\toprule
Method & Backbone & HOTA$\uparrow$ & MOTA$\uparrow$ & IDF1$\uparrow$ & FPS$\uparrow$ \\
\midrule
MOTR & ResNet-50 & 54.2 & 79.7 & 51.5 & 9.5\\
MOTRv2 & ResNet-50 & 69.9 & 91.9 & 71.7 & 6.9\\
MOTRv3 & ResNet-50 & 68.3 & 91.7 & 70.1 & \bf 10.6\\
\midrule
MOTRv3 & ConvNeXt-B & \bf 70.4 & \bf 92.9 & \bf 72.3 & 9.8 \\
\bottomrule
\end{tabular}
}
\label{table:fps}
\end{center}
\vspace{-3mm}
\end{table}


\end{document}